# Understanding Semantic Web and Ontologies:
# Theory and Applications

Mohammad Mustafa Taye
Department of Software Engineering
Faculty of Information Technology
Philadelphia University,
Amman, Jordan

**Abstract —** Semantic Web is actually an extension of the current one in that it represents information more meaningfully for humans and computers alike. It enables the description of contents and services in machine-readable form, and enables annotating, discovering, publishing, advertising and composing services to be automated. It was developed based on Ontology, which is considered as the backbone of the Semantic Web. In other words, the current Web is transformed from being machine-readable to machine-understandable. In fact, Ontology is a key technique with which to annotate semantics and provide a common, comprehensible foundation for resources on the Semantic Web. Moreover, Ontology can provide a common vocabulary, a grammar for publishing data, and can supply a semantic description of data which can be used to preserve the Ontologies and keep them ready for inference. This paper provides basic concepts of web services and the Semantic Web, defines the structure and the main applications of ontology, and provides many relevant terms are explained in order to provide a basic understanding of ontologies.

## 1 INTRODUCTION

One of the most interesting inventions, in recent decades, is that of Web Services [36]. These are computer program "applications": self-describing, self-contained applications whose function is to automatically share information over the Internet with other applications. Some weaknesses such as browsing information without taking its meaning into account have recently appeared in Web Services. This creates a need for a new Web with more relevance to the user.

Semantic Web is actually an extension of the current one in that it represents information more meaningfully for humans and computers alike. It enables the description of contents and services in machine-readable form, and enables annotating, discovering, publishing, advertising and composing services to be automated. It was developed based on Ontology, which is considered as the backbone of the Semantic Web. In other words, the current Web is transformed from being machine-readable to machine-understandable.

One function of the Web is to build a source of reference for information on several subjects, while the Semantic Web is designed to build a web of meaning. The foundation of vocabularies and effective communication on the Semantic Web is ontology. "Ontology provides a formal, explicit specification of a shared conceptualisation of a domain" [31, 36]. Therefore, it facilitates knowledge sharing over distributed systems; in other words, it allows systems or applications to cooperate that were not formerly designed to interoperate. Ontology plays a major part in solving the problem of interoperability between applications across different organizations, by providing a shared understanding of common domains.

Several Ontologies have recently been built. Consequently, they should be accessed from other applications for use or information exchange. Ontologies in such numbers present interoperability problems, for which many solutions have been developed. One of these is to build a single Ontology, but this is inadequate, partly because it is too inflexible for knowledge sharing. Another solution is Ontology Mapping which plays an important role in solving interoperability in heterogeneous systems and in many application domains. This way is to build bridges between Ontologies in order to provide commonly accessible layers that could then exchange information in semantically sound ways. Therefore, in the following sections, this paper will describe many terminologies in order to understand the Semantic Web and Ontologies.

## 2. WEB SERVICE

A web service [36] is a self-describing software program using self-contained applications and identified by a Uniform Resource Identifier (URI), used to share information between applications over the Internet. Access to and retrieval of information from the Web occurs via the HTTP protocol. One of the first languages to have been used for the internet is HTML, a markup language used to describe the document structure. The Web can be conceived as a huge library containing a large amount of information or data – unfortunately without any sensible means of representation.

The common Web service scenario [36] can be described by the three actions of publish, bind and find, and three actors: the service requester, the service provider and the registry, where services can be published, advertised and sometimes located. In other words, service providers describe and advertise their services in the registry, while service requesters search the registry for services that match their requirements. There are obviously many examples of Web services, including:

- Credit card authorisation.
- Currency converter (e.g. dollars to Euros).
- Stock quote provider.
- Shipping rate calculator.

## 3 SEMANTIC WEB

The Semantic Web [36] is distributed and heterogeneous, has brought the evolution of the Web to a higher level. There are two visions of the future in the development of the Web, the first being to improve its usability as a medium for collaboration and the second to ensure that its contents can be understood by machines. Providing annotation data will facilitate this second aim.

Tim Berners-Lee, who invented the WWW and has worked on the Semantic Web, states that the latter "is not a separate Web but an extension of the current one, in which information is given a well-defined meaning, better enabling computers and people to work in cooperation." [2]. Thus, the Semantic Web [16, 31] is distinguished by a more meaningful representation of information for humans and computers, providing a description of its contents and services in machine-readable form; moreover, it enables services to be automatically annotated, discovered, published, advertised and composed. It thereby facilitates interoperability and the sharing of knowledge over the Web. Its main goal is therefore to make information on the Web accessible and understandable by humans and computers. Figure 1 illustrates the architecture of the Semantic Web.

In fact, both the Semantic Web and Web services are considered to be a set of resources, identified by the URI. The difference between them is that Web services use HTTP to display the contents of a page, while the Semantic Web tries to create machine readability by semantically representing the data or information in resources. Numerous tools and applications of Semantic Web technologies have recently become available.

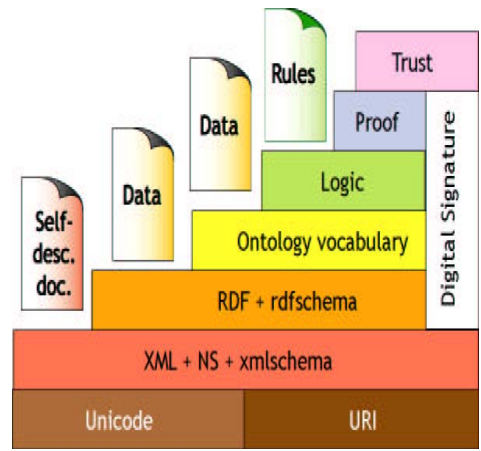

Figure 1: Semantic Web architecture [2]

The layers of architecture represented [2, 16, 31, 36, 35] in Figure 1 are briefly described below:

- **URI and Unicode:** To identify and locate resources, or indeed anything on the Web, a uniform system of identifiers (URIs) is used. The URI, which is considered to be the foundation of the Web, is used to give a unique name to each resource. Unicode is the standard for computer character representation.
- **Extensible Markup Language (XML)** is a markup language, which means that it is machine-readable and has its own format. It is widely known in the WWW community because it has a flexible text format and was designed to describe data and to meet the challenges of large-scale e-business and electronic publishing; it plays an important role in exchanging different types of data on the Web. In fact, it is the basis of a rapidly growing number of software development activities. Each document starts with a namespace declaration using XML Namespace.
- **The Resource Description Framework (RDF)** is the first layer of the Semantic Web. RDF is a framework for using and representing metadata and describing the semantics of information about resources on the Web in a machine-accessible way. It uses URIs to identify Web resources and to describe the relations between these resources, using a graph model. While describing classes of resources and the properties between them, using RDF Schema (which is a simple modelling language), it also provides a simple reasoning framework for inferring types of resources.
- **Ontology Vocabulary** is a language which provides a common vocabulary and grammar for published data as well as a semantic description of the data used to preserve the ontologies and to keep them ready for inference. Ontology means describing the semantics of the data, providing a uniform way to enable communication by which different parties can understand each other.
- **Logic and Proof:** In the Semantic Web, the building of systems follows a logic which considers the structure of ontology. A reasoner could be used to check and resolve consistency problems and the redundancy of the concept translation. A reasoning system is used to make new inferences.
- **Trust** is the final layer of the Semantic Web. This component concerns the trustworthiness of the information on the Web in order to provide an assurance of its quality.

## 4 ONTOLOGY

Ontologies [27], which are used in order to support interoperability and common understanding between the different parties, are a key component in solving the problem of semantic heterogeneity, thus enabling semantic interoperability between different web applications and services.

Recently, ontologies have become a popular research topic in many communities, including knowledge engineering, electronic commerce, knowledge management and natural language processing. Ontologies provide a common understanding of a domain that can be communicated between people, and of heterogeneous and widely spread application systems. In fact, they have been developed in Artificial Intelligence (AI) research communities to facilitate knowledge sharing and reuse.

The goal of an ontology is to achieve a common and shared knowledge that can be transmitted between people and between application systems. Thus, ontologies [16] play an important role in achieving interoperability across organizations and on the Semantic Web [36], because they aim to capture domain knowledge and their role is to create semantics explicitly in a generic way, providing the basis for agreement within a domain. Ontology is used to enable interoperation between Web applications from different areas or from different views on one area. For that reason, it is necessary to establish mappings among concepts of different ontologies to capture the semantic correspondence between them. However, establishing such a correspondence is not an easy task [31].

Because there are many different definitions of ontology, it is very difficult to find a definition that researchers can agree upon. The present research first presents some of these definitions which have been given from different perspectives, and then explores in depth those aspects of these definitions which are related to the topic under investigation.

The primary use of the word "ontology" is in the discipline of philosophy, where it means "the study or theory of the explanation of being"; it thus defines an entity or being and its relationship with and activity in its environment. In other disciplines, such as software engineering and AI, it is defined as "a formal explicit specification of a shared conceptualization" [12]. The foundations of this definition are:

- **All knowledge** (e.g. the type of concepts used and the constraints on their use) in ontology must have an explicit specification.
- An ontology is a **conceptualisation**, which means it has a universally comprehensible concept.
- "**Shared**" indicates an agreement about the meaning in such domains. In other words, an ontology should capture consensual knowledge accepted by the communities.
- "**Formal**" refers to the grounding of representation in well understood logic, and any ontology should be machine-processable.

### 4.1 ONTOLOGY REPRESENTATION

Ontology is comprised of four main components: concepts, instances, relations and axioms. The present research adopts the following definitions of these ontological components:

- **A Concept** (also known as a class or a term) is an abstract group, set or collection of objects. It is the fundamental element of the domain and usually represents a group or class whose members share common properties. This component is represented in hierarchical graphs, such that it looks similar to object-oriented systems. The concept is represented by a "super-class", representing the higher class or so-called "parent class", and a "subclass" which represents the subordinate or so-called "child class". For instance, a university could be represented as a class with many subclasses, such as faculties, libraries and employees.
- **An Instance** (also known as an individual) is the "ground-level" component of an ontology which represents a specific object or element of a concept or class. For example,

"Jordan" could be an instance of the class "Arab countries" or simply "countries".

- **A Relation** (also known as a slot) is used to express relationships between two concepts in a given domain. More specifically, it describes the relationship between the first concept, represented in the domain, and the second, represented in the range. For example, "study" could be represented as a relationship between the concept "person" (which is a concept in the domain) and "university" or "college" (which is a concept in the range).
- **An Axiom** is used to impose constraints on the values of classes or instances, so axioms are generally expressed using logic-based languages such as first-order logic; they are used to verify the consistency of the ontology.

### 4.2 STRUCTURE OF ONTOLOGY

In general, the structure of an ontology [28] is described as a

5-*tuple O:* = (C, $H^C$, R, $H^R$, I),

where

- C represents a set of concepts (instances of "*rdf:Class*"). These concepts are arranged with a corresponding subsumption hierarchy $H^C$.

- R represents a set of relations that relate concepts to one another (instances of "*rdf:Property*"). $R_i \in R$ and $R_i \rightarrow C \times C$.

- $H^C$ represents a concept hierarchy in the form of a relation (a binary relation corresponding to "*rdfs:subClassOf*"). $H^C \subseteq C \times C$, where $H^C (C_1, C_2)$ denotes that $C_1$ is a subconcept of $C_2$.

- $H^R$ represents a relation hierarchy in the form of a relation $H^R \subseteq R \times R$, where $H^R (R_1, R_2)$ denotes that $R_1$ is a subrelation of $R_2$("*rdfs:subPropertyOf*").

- I is the instantiation of the concepts in a particular domain ("*rdf:type*").

In general, there are many ways to represent or model the classification of concepts semantically. These include taxonomies, thesauri and ontologies. These widely varying concepts are used in Web semantics, which is why it is necessary to apply taxonomy, or the science of identifying and arranging vocabulary in the shape of a hierarchy or tree. In other words, it is used to describe concepts and their relationships explicitly. The relationships of "subclass" and "super-class" are the taxonomic ones that could be used.

A thesaurus [26] is a vocabulary with conceptual relationships between the terms and can be considered an extension of taxonomy with richer semantic relationships. It can easily be converted into a taxonomy, but expressiveness and semantics will be lost. The relationships which could be used in a thesaurus are equivalence, homography, hierarchy and association.

Ontologies are like taxonomies but with more semantic relationships between concepts and attributes; they also contain strict rules used to represent concepts and relationships. An ontology is a hierarchically structured set of terms for describing a domain that can be used as a skeletal foundation for a knowledge base. According to this definition, the same ontology can be used for building several knowledge bases, which would share the same skeleton or taxonomy.

The ontology community distinguishes ontologies that are mainly taxonomies from those that model the domain in a deeper way and provide more restrictions on domain semantics. The community calls them lightweight and heavyweight ontologies respectively. The former include concepts, concept taxonomies, relationships between concepts, and properties that describe concepts, while heavyweight ontologies add axioms and constraints to lightweight ones. These clarify the intended meaning of the terms gathered in the ontology.

Heavyweight and lightweight ontologies can be modelled with different knowledge modelling techniques and can be implemented in various kinds of languages [21]. Ontologies can be:

- **highly informal** if they are expressed in natural language; According to this, a highly informal ontology would not be an ontology, since it is not machine-readable.
- **semi-informal** if expressed in a restricted and structured form of natural language, since it is a machine-readable;
- **semi-formal** if expressed in an artificial and formally defined language (e.g. RDF graphs); and
- **rigorously formal** if they provide meticulously defined terms with formal semantics, theorems and proof of properties such as soundness and completeness (e.g. Web Ontology Language [OWL]).

The expressiveness of an ontology is based on the degree of explication of the (meta-) knowledge. Several ontologies capture specific domains or certain applications, while others try to capture all terms in natural language. Ontologies that capture extra relations and extra constraints are considered to be more expressive, because they capture knowledge of the domain on a more detailed level. On the other hand, the expressiveness of an ontology is restricted by the languages used for describing or specifying it. Ontology languages can be seen as restricting the expressiveness of the ontology [38].

An ontology is expressed in a knowledge representation language, which provides a formal frame of semantics. This ensures that the specification of domain knowledge in an ontology is machine-processable and is being interpreted in a well-defined way [2].

## 4.3 ONTOLOGY APPLICATIONS

Over the years, ontology has become a popular research topic in a range of disciplines, with the aim of increasing understanding of and building a consensus in a given area of knowledge. Ontology also leads to the sharing of knowledge between systems and people. As mentioned previously, ontology first appeared in AI laboratories, before being used in other fields; for example:

- **Semantic Web** [2]: Ontology plays a key role in the Semantic Web in supporting information exchange across distributed environments. The Semantic Web represents data in a machine-processable way, which is why it is considered to be an extension of the current Web.
- **Semantic Web Service Discovery** [2]: In the e-business environment, ontology plays an important role by finding the best match for the requester looking for merchandise or something else. It also helps online travel customers obtain a response.
- **Artificial Intelligence** [27]: Ontology has been developed in the AI research community, its goal here being to facilitate the sharing of knowledge and the reuse and enabling of processing between programs, services, agents or organisations across a given domain.
- **Multi-agent** [16]: The importance of ontology in this area is that it provides a shared understanding of domain knowledge, allowing for easy communication between agents and thereby reducing misunderstandings.
- **Search Engines** [16, 31]: These use ontology in the form of thesauri to find the synonyms of search terms, which facilitates internet searching.
- **E-Commerce** [16]: This application uses ontology to facilitate communication between seller and buyer through the description of merchandise, as well as enabling machine-based communication.

- **Interoperability** [7]: The problem of bringing together heterogeneous and distributed systems is known as the "interoperability problem". In this area, the importance of applying ontology appears explicitly: it is used to integrate different heterogeneous application systems.

In the field of services, ontology plays the major role of providing a richer description of these services and terms and the relationships between them in the application domain, leading to a capture of the domain of knowledge in an explicitly representative manner. At the same time, it supports the inference of implied knowledge by declaring the descriptions.

The following example is given in order to demonstrate the reasons for considering ontology to be the backbone of the Semantic Web. As mentioned in [25], it illustrates how ontology may be used to match services with semantic meanings. According to this scenario, the service requester invokes a service (for example, buying a car), which triggers a description of the service request information annotated in metadata. Service providers also describe and advertise their services in metadata to provide answers to the requester, while the service match engine receives the metadata of both provider and requester, upon which it accesses the ontology, which provides a possible identification of "automobile" and "vehicle" with "car". The service match engine will infer from this whether the request has been satisfied or not (see Figure 2).

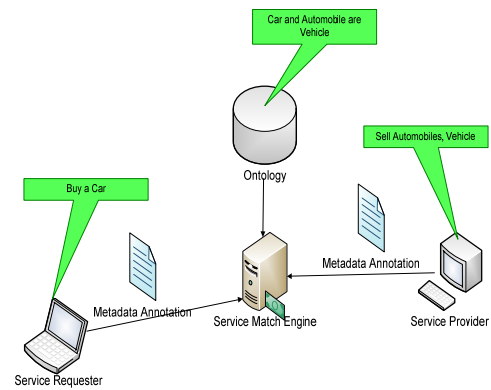

Figure 2: Ontology in service over the internet

## 4.4 ONTOLOGY INTEROPERABILITY

This section describes several operations on ontologies.

### 4.4.1 ONTOLOGY TRANSFORMATION AND TRANSLATION

Ontology Transformation [6, 9] is the process used to develop a new ontology to cope with new requirements made by an existing one for a new purpose, by using a transformation function t. In this operation, many changes are possible, including changes in the semantics of the ontology and changes in the representation formalism.

Ontology Translation is the function of translating the representation formalism of an ontology while keeping the same semantic. In other words, it is the process of change or modification of the structure of an ontology in order to make it suitable for purposes other than the original one.

There are two types of translation. The first is translation from one formal language to another, for example from RDFS to OWL, called syntactic translation. The second is translation of vocabularies, called semantic translation [6].

The translation problem arises when two Web-based agents attempt to exchange information, describing it using different ontologies.

### 4.4.2 ONTOLOGY MERGING

Ontology merging [20, 23, 30] is the process of creating a new single coherent ontology from two or more existing source ontologies related to the same domain. The new ontology will replace the source ontologies.

### 4.4.3 ONTOLOGY INTEGRATION

Integration [23, 30] is the process of creating a new ontology from two or more source ontologies from different domains.

### 4.4.4 ONTOLOGY MAPPING

Ontology mapping [19, 22, 29, 32, 34] is a formal expression or process that defines the semantic relationships between entities from different ontologies. In other words, it is an important operator in many ontology application domains, such as the Semantic Web and e-commerce, which are used to describe how to connect and from correspondences between entities across different ontologies. Ontology matching is the process of discovering similarities between two ontologies.

An entity e is understood in an ontology O denoted by e|O is concept C, relation R, or instance I, i.e. $e|O \in C \cup R \cup I$. Mapping the two ontologies, O1 onto O2, means that each entity in ontology O1 is trying to find a corresponding entity which has the same intended meaning in ontology O2.

The Ontology mapping function "map" is defined based on the vocabulary, E, of all terms $e \in E$ and based on the set of possible ontologies, O as a partial function:

$$\text{map}: E \times O \times O \to E, \text{ with}$$

$$\forall e \in O1 (\exists f \in O2 : \text{map}(e, O1, O2) = f \vee \text{map}(e, O1, O2) = \bot).$$

An entity is mapped to another entity or none.

### 4.4.5 ONTOLOGY ALIGNMENT

Ontology alignment [12, 13, 14, 24] is the process or method of creating a consistent and coherent link between two or more ontologies by bringing them into mutual agreement. This method is near to artificial intelligence methods: being a logical relation, ontology alignments are used to clearly describe how the concepts in the different ontologies are logically related. This means that additional axioms describe the relationship between the concepts in different ontologies without changing the meaning in the original ontologies.

In fact, ontology alignment uses as pre-process for both ontology merging and ontology integration.

There are many different definitions of ontology alignment, depending upon its application and its intended outcome. Sample definitions include the following:

- Ontology alignment is used to "establish correspondences among the source ontologies, and to determine the set of overlapping concepts, concepts that are similar in meaning but have different names or structure, and concepts that are unique to each of the sources" [30].
- Ontology alignment is the process of bringing two or more ontologies into mutual agreement, making them consistent and coherent [7, 10, 33].
- "Given two ontologies O1 and O2, mapping one ontology onto another means that each entity (concept C, relation R, or instance I) in ontology O1 is trying to find a corresponding entity (i.e. by using matching algorithms), which has the same intended meaning, in ontology O2" [11].

Formally, an ontology alignment function is defined as follows:

An ontology alignment function, align, based on the set E of all entities $e \in E$ and based on the set of possible ontologies O, is a partial function.

$$\text{Align}: O1 \to O2$$

$$\text{Align}(e_{O1}) = f_{O2} \text{ if } \text{Sim}(e_{O1}, f_{O2}) > \text{threshold}.$$

Where $O_i$: ontology, $e_{Oi}$, $f_{Oj}$: entities of ($O_i$, $O_j$) Sim ($e_{O1}$, $f_{O2}$): similarity function between two entities $e_{O1}$ and $f_{O2}$.

The ontology alignment function is based on different similarity measures.

A similarity measure is a real-valued function Sim ($e_i$, $f_j$): $O \times O \rightarrow [0, 1]$ measuring the degree of similarity between x and y.

$$Sim(e_i, e_j) = \begin{cases} 1 & e_i = e_j \quad \text{two entities are identical} \\ 0 & e_i \neq e_j \quad \text{two entities are different} \end{cases}$$

### 4.4.6 OTHER ONTOLOGY OPERATIONS

There are many operations that could apply to ontology, such as changing [23], which is considered one of the most interesting and important operations that should be taken into account when dealing with ontology. In general, most existing ontologies have large sizes and complex structures. In fact, several factors could be responsible for a change in ontology, including a response to new needs or requirements, a change by the developer or the editor of ontology, an ontological translation from one language to another and a change of domain of interest. On the other hand, using versioning could help to reduce those problems by keeping track of the relationships between different revisions of ontology [23]. As argued in [23] ontology versioning is the ability to handle changes in ontologies by creating and managing different ontological variants.

### 5. ONTOLOGY LANGUAGES, ONTOLOGY EDITORS AND DEVELOPMENT TOOLS

The main object of semantic web languages [35] is to add semantics to the existing information on the Web. RDF/RDFS [5], OIL [15], DAML+OIL [18] and OWL [1] are modelling web languages that have been developed to represent or express ontologies. In general, most of them are based on XML [4] syntax, but they have different terminologies and expressions. Indeed, some of these languages have the ability to represent certain logical relations which others do not. Because some languages have greater expressive power than others, their selection for representing ontologies is based mainly on what the ontology represents or what it will be used for. In other words, different kinds of ontological knowledge-based applications need different language facilitators to enable reasoning on ontology data. These description languages provide richer constructors for forming complex class expressions and axioms.

In fact, most recent ontology developers have used ontology editors, which are environments or tools used directly for editing, developing or modifying ontologies. They are used to provide support to the ontological development process, as well as to conceptualise the ontology; they transform the conceptualisation into an executable code using translators. Therefore, the output ontology of these tools will be in one of the Web ontology languages supported by editors such as Protégé [30], OWL-P [8] and OilEd [3]. Alternatively, ontology reasoners are used to check the conflicts with a high degree of automation. Many such systems have recently been developed, including RACER [17] and FaCT [37].

Returning to the main concern of this section, modelling web languages, there are in general two different types: presentation languages such as HTML, designed to represent text and images to users or requesters without reference to the content, and data languages, intended to be processed by machines. The present research relates to the latter.

Before OWL, much research had been conducted into creating a powerful ontology modelling language. This research stream began with the XML-based RDF and RDF/S, progressed to the Ontology Inference Layer (OIL) and continued with the creation of DAML+OIL, the result of joining the American proposal DAML-ONT5 with the European language OIL. All these languages are based on XML or RDF syntax and are consequently compatible with web standards. Indeed, RDF and OWL make searching for and reusing information both easier and more reliable, because they are considered as standards that enable the Web to be a global infrastructure for sharing documents and data equally.

As mentioned in [1, 35], some important requirements for quality support should be taken into account when developing languages for encoding ontologies. These include giving the user explicit written format, ease of use, expressive power, compatibility, sharing and versioning, internationalisation, formal conceptualisations of domain models, well-defined syntax and semantics, efficient reasoning support, sufficient expressive power and convenience of expression.

Syntax is one of the most important features of any language, so it should be well-defined; it is also the most significant condition required for the processing of information by machine.

The semantics of knowledge should be well defined, because it represents the meaning of that knowledge. Formal semantics should be established in the domain of mathematical logic in a clearly defined way that will lead to unambiguous meaning, since well defined semantics will lead to correct reasoning. Semantics can be considered a prerequisite to support reasoning. On the other hand, reasoning will help to check and discover consistent ontology, to verify unintended relationships between classes and to classify individuals into classes.

This section has detailed the most common and important languages, RDF, RDF/S, DAML+OIL and OWL, all of which are based on XML. XML itself [4] is widely known in the WWW community, because it is a flexible text format designed to describe data and to meet the challenges of large-scale e-business and electronic publishing, which plays an important role in exchanging different types of data on the Web. In fact, it is the basis of a rapidly growing number of software development activities.

## 8. CONCLUSION

The goal of an ontology is to achieve a common and shared knowledge that can be transmitted between people and between application systems. Thus, ontologies play an important role in achieving interoperability across organizations and on the Semantic Web, because they aim to capture domain knowledge and their role is to create semantics explicitly in a generic way, providing the basis for agreement within a domain. Thus, ontologies have become a popular research topic in many communities. In fact, ontology is a main component of this research; therefore, the definition, structure and the main operations and applications of ontology are provided.